\documentclass[10pt,twocolumn,letterpaper]{article}

\usepackage[pagenumbers]{cvpr} 

\usepackage{graphicx}
\usepackage{amsmath}
\usepackage{amssymb}
\usepackage{booktabs}
\usepackage[utf8]{inputenc} 
\usepackage[T1]{fontenc}    
\usepackage{url}            
\usepackage{booktabs}       
\usepackage{amsfonts}       
\usepackage{nicefrac}       
\usepackage{microtype}      
\usepackage{xcolor}         
\usepackage{amsmath,amssymb}
\usepackage{graphicx}
\usepackage{mathtools}
\usepackage{algorithm}
\usepackage{algorithmic}
\usepackage{setspace}
\usepackage{amsthm}
\usepackage{booktabs}  
\usepackage{arydshln}
\usepackage{multirow}
\usepackage{url}
\usepackage{graphicx} 
\usepackage{graphics} 
\usepackage{booktabs} 
\usepackage{multirow} 
\usepackage{mathrsfs} 
\usepackage{amsmath} 
\usepackage{booktabs} 
\usepackage{bm}

\usepackage{tikz}

\tikzstyle{mathtext}=[text badly centered, font={\fontsize{8pt}{8pt}\selectfont}]
\tikzstyle{smalltext}=[text badly centered, font={\fontsize{5pt}{5pt}\selectfont}]

\usepackage{amsmath,amsfonts, bm}

\def\1{\bm{1}}

\def\vepsilon{{\bm{\epsilon}}}

\def\vg{{\bm{g}}}

\def\vx{{\bm{x}}}

\def\mI{{\bm{I}}}

\DeclareMathAlphabet{\mathsfit}{\encodingdefault}{\sfdefault}{m}{sl}
\SetMathAlphabet{\mathsfit}{bold}{\encodingdefault}{\sfdefault}{bx}{n}

\def\gL{{\mathcal{L}}}

\def\gN{{\mathcal{N}}}

\DeclarePairedDelimiterX{\infdivx}[2]{(}{)}{%
    #1\;\delimsize\|\;#2%
}
\newcommand{\kl}{D_{\mathrm{KL}}\infdivx}

\newcommand{\E}{\mathbb{E}}

\definecolor{cvprblue}{rgb}{0.21,0.49,0.74}
\usepackage[pagebackref,breaklinks,colorlinks,citecolor=cvprblue]{hyperref}

\newcommand\blfootnote[1]{%
\begingroup 
\renewcommand\thefootnote{}\footnote{#1}%
\addtocounter{footnote}{-1}%
\endgroup 
}

\usepackage[capitalize]{cleveref}
\crefname{section}{Sec.}{Secs.}
\Crefname{section}{Section}{Sections}
\Crefname{table}{Table}{Tables}
\crefname{table}{Tab.}{Tabs.}

\begin{document}
 
\title{LucidDreamer: Towards High-Fidelity Text-to-3D Generation\\ via Interval Score Matching}

\author{
Yixun Liang$^{\text{\textbf{\textcolor{red}{*}}}1}$ ~
Xin Yang$^{\text{\textbf{\textcolor{red}{*}}}1, 2}$~
Jiantao Lin$^{1}$ ~
Haodong Li$^{1}$ ~
Xiaogang Xu$^{3, 4}$ ~ 
Yingcong Chen$^{** 1, 2}$
\\
$^1$ HKUST (GZ) \quad
$^2$ HKUST  \quad 
$^3$ Zhejiang Lab \quad $^4$ Zhejiang University
\\
 {\tt \small yliang982@connect.hkust-gz.edu.cn} \quad  {\tt \small xin.yang@connect.ust.hk}  \quad  {\tt \small jlin695@hkust-gz.edu.cn} 
 \\
 {\tt \small hli736@connect.hkust-gz.edu.cn} \quad {\tt \small xgxu@zhejianglab.com}  \quad {\tt \small yingcongchen@ust.hk}
}

\twocolumn[{
\maketitle
\begin{figure}[H]
    \vspace{-2em}
    \hsize=\textwidth
    \centering
    \includegraphics[width=2\linewidth]{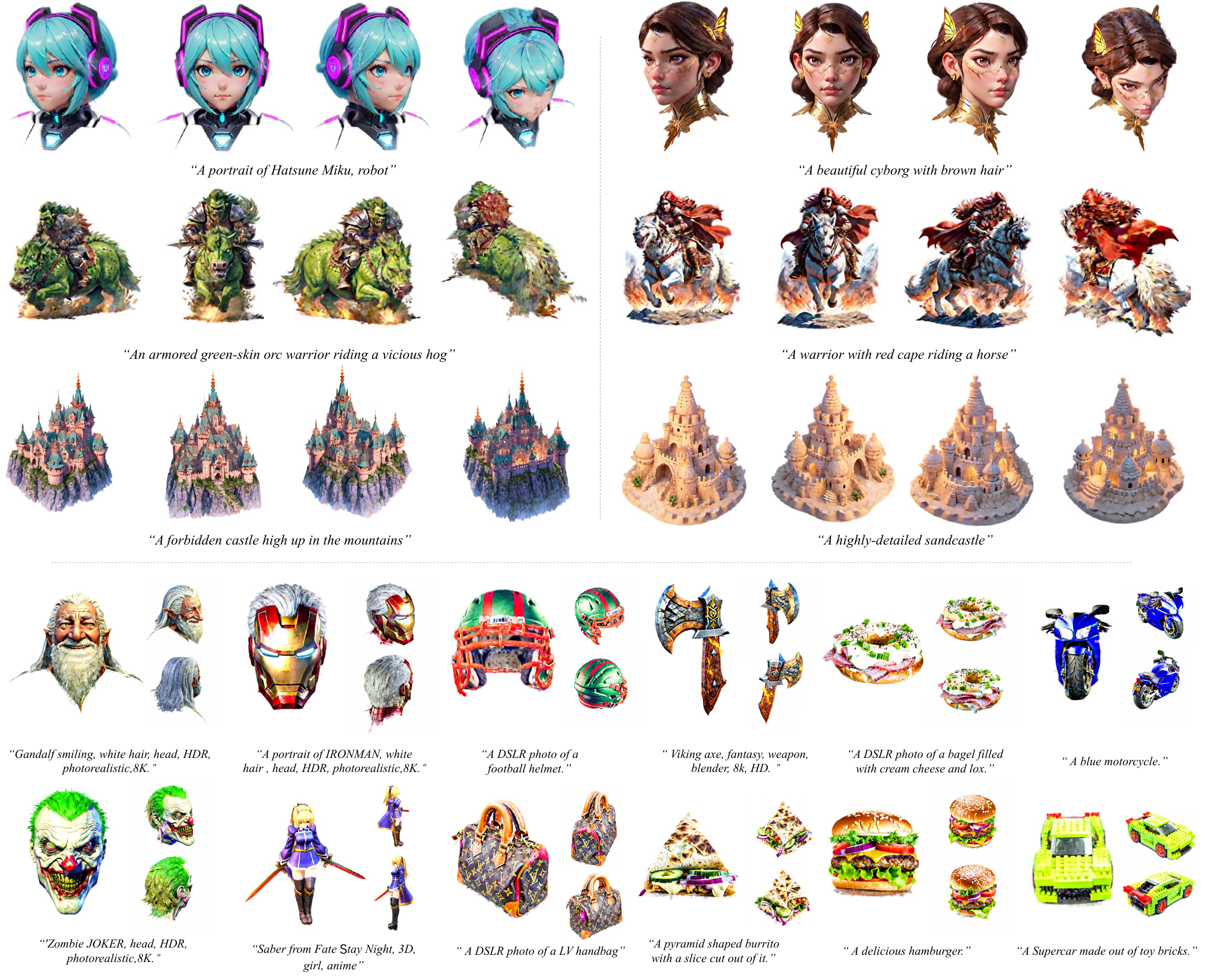}
    \vspace{-0.5em}
    \caption{\textbf{Examples of text-to-3D content creations with our framework.} We present a text-to-3D generation framework, named the \textit{LucidDreamer}, to distill high-fidelity textures and shapes from pretrained 2D diffusion models (detailed shows on Sec.~\ref{sec:experiments}) with a novel \textbf{Interval Score Matching} objective and an \textit{Advanced 3D distillation pipeline}. Together, we achieve superior 3D generation results with photorealistic quality in a short training time. Please zoom in for details.}
    \label{fig:teaser}
    \vspace{-1em}
\end{figure}
}]

\blfootnote{** Corresponding author.}
\blfootnote{\textcolor{red}{*The first two authors contributed equally to this work.}} 

\blfootnote{\textcolor{red}{*} \textit{Conceptualization}: \textbf{Yixun Liang}: 60\%, \textbf{Xin Yang}: 40\%,}
\blfootnote{\textcolor{white}{*}  \textit{Methodology}: ~~~~~~~~\textbf{Xin Yang}: 60\%, \textbf{Yixun Liang}: 40\%.}

\begin{abstract}
The recent advancements in text-to-3D generation mark a significant milestone in generative models, unlocking new possibilities for creating imaginative 3D assets across various real-world scenarios. While recent advancements in text-to-3D generation have shown promise, they often fall short in rendering detailed and high-quality 3D models. This problem is especially prevalent as many methods base themselves on Score Distillation Sampling (SDS). This paper identifies a notable deficiency in SDS, that it brings inconsistent and low-quality updating direction for the 3D model, causing the over-smoothing effect. To address this, we propose a novel approach called Interval Score Matching (ISM). ISM employs deterministic diffusing trajectories and utilizes interval-based score matching to counteract over-smoothing. Furthermore, we incorporate 3D Gaussian Splatting into our text-to-3D generation pipeline. Extensive experiments show that our model largely outperforms the state-of-the-art in quality and training efficiency. Our code will be available at: \href{https://github.com/EnVision-Research/LucidDreamer}{EnVision-Research/LucidDreamer}
\end{abstract}

\section{Introduction}
\label{sec.intro}
Digital 3D asserts have become indispensable in our digital age, enabling the visualization, comprehension, and interaction with complex objects and environments that mirror our real-life experiences. Their impact spans a wide range of domains including architecture, animation, gaming, virtual and augmented reality, and is widely used in retail, online conferencing, education, etc. The extensive use of 3D technologies brings a significant challenge, i.e., generating high-quality 3D content is a process that needs a lot of time, effort, and skilled expertise.

This stimulates the rapid developments of 3D content generation approaches~\cite{poole2022dreamfusion,lin2023magic3d,michel2022text2mesh,chen2023fantasia3d,wang2023prolificdreamer,Liu_Wu_Hoorick_Tokmakov_Zakharov_Vondrick_2023,Hong_Zhang_Pan_Cai_Yang_Liu_2022,Lin_Han_Gong_Xu_Zhang_Li_2023,liu2023one,huang2023dreamwaltz,purushwalkam2023conrad,shi2023MVD,Metzer_Richardson_Patashnik_Giryes_Cohen}. Among them, text-to-3D generation~\cite{poole2022dreamfusion, michel2022text2mesh,chen2023fantasia3d,wang2023prolificdreamer,zhu2023hifa,Hong_Zhang_Pan_Cai_Yang_Liu_2022,lin2023magic3d, Metzer_Richardson_Patashnik_Giryes_Cohen} stands out for its ability to create imaginative 3D models from mere text descriptions. This is achieved by utilizing a pretrained text-to-image diffusion model as a strong image prior to supervise the training of a neural parameterized 3D model, enabling for rendering 3D consistent images in alignment with the text. This remarkable capability is fundamentally grounded in the use of Score Distillation Sampling (SDS). SDS acts as the core mechanism that lifts 2D results from diffusion models to the 3D world, enabling the training of 3D models without images ~\cite{poole2022dreamfusion,chen2023fantasia3d, lin2023magic3d, Zhang_Chen_Yang_Qu_Wang_Chen_Long_Zhu_Du_Zheng_2023, Cao_Cao_Han_Shan_Wong_2023,huang2023dreamwaltz, Metzer_Richardson_Patashnik_Giryes_Cohen}. 

\begin{figure}[t]
    \includegraphics[width=1.0\linewidth]{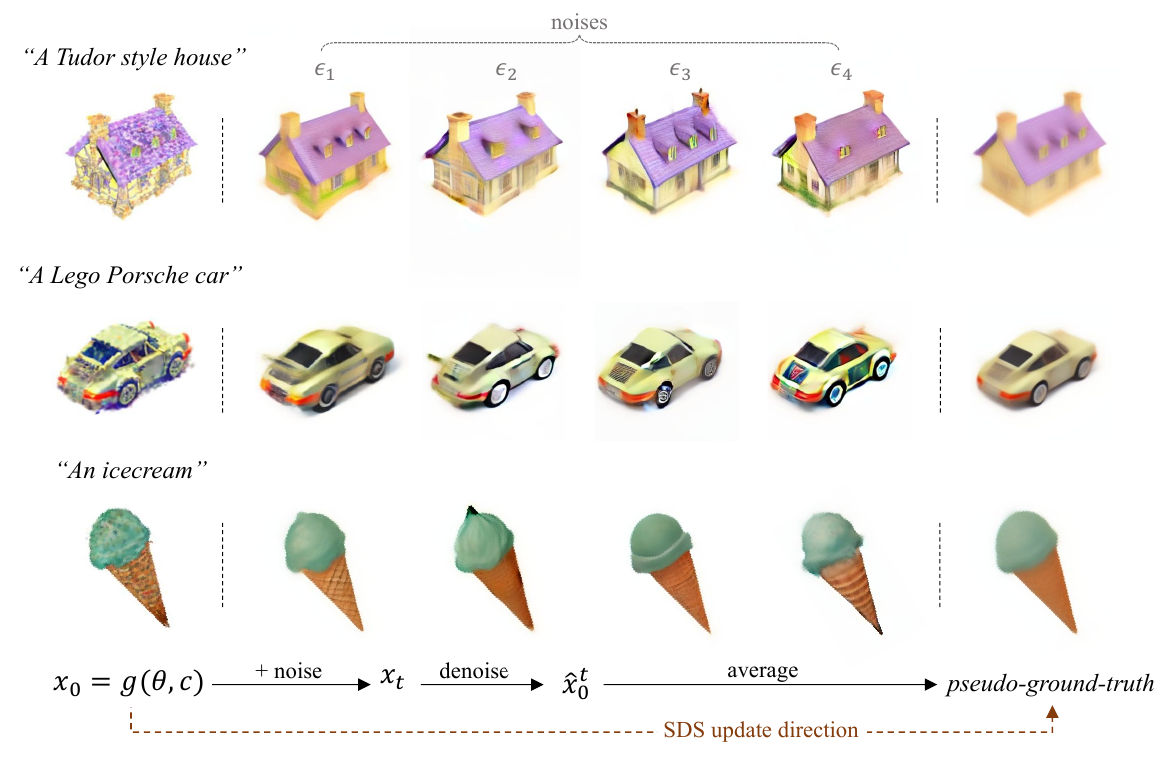}
    \vspace{-1.5em}
    \caption{\textbf{Examples of SDS~\cite{poole2022dreamfusion}}. Let $t = 500$, we simulate the SDS distillation process by sampling $x_t$ with same $x_0$ but different noises $\{\vepsilon_1, ..., \vepsilon_4\}$. We discover that the SDS distillation process produces overly-smoothed \textit{pseudo-ground-truth} (i.e., $\hat{x}_0^t$) for $x_0$. First, the random noise and timestep sampling strategy of SDS drives $x_0$ towards the averaged $\hat{x}_0^t$ and eventually leads to the ``feature-averaging'' result. Second, SDS exploits the diffusion model for $\hat{x}_0^t$ estimation in one step, which results in low-quality guidance at large timesteps. Please refer to Sec.~\ref{sec:sds_revisit} for more analysis.}
    \label{fig:sds_drawbacks}
    \vspace{-0.8em}
\end{figure}

Despite its popularity, empirical observations have shown that SDS often encounters issues such as over-smoothing, which significantly hampers the practical application of high-fidelity 3D generation. In this paper, we thoroughly investigate the underlying cause of this problem. Specifically, we reveal that the mechanism behind SDS is to match the images rendered by the 3D model with the pseudo-Ground-Truth (pseudo-GT) generated by the diffusion model. However, as shown in Fig.~\ref{fig:sds_drawbacks}, the generated pseudo-GTs are usually \textit{inconsistent} and have \textit{low visual quality}. Consequently, all update directions provided by these pseudo-GTs are subsequently applied to the same 3D model. Due to the average effect, the final results tend to be over-smooth and lack of details. 

This paper aims to overcome the aforementioned limitations. We show that the unsatisfactory pseudo-GTs originated from two aspects. Firstly, these pseudo-GTs are one-step reconstruction results from the diffusion models, which have high reconstruction errors. Besides, the intrinsic randomness in the diffusion trajectory makes these pseudo-GTs semantically variant, which causes an averaging effect and eventually leads to over-smoothing results. To address these issues, we propose a novel approach called Interval Score Matching (ISM). ISM improves SDS with two effective mechanisms. Firstly, by employing DDIM inversion, ISM produces an invertible diffusion trajectory and mitigates the averaging effect caused by pseudo-GT inconsistency. 
Secondly, rather than matching the pseudo-GTs with images rendered by the 3D model, ISM conducts matching between two interval steps in the diffusion trajectory, which avoids one-step reconstruction that yields high reconstruction error. 
We show that our ISM loss consistently outperforms SDS by a large margin with highly realistic and detailed results. Finally, we also show that our ISM is not only compatible with the original 3D model introduced in \cite{poole2022dreamfusion}, by utilizing a more advanced model -- 3D Gaussian Splatting \cite{kerbl3Dgaussians}, our model achieves superior results compared to the state-of-the-art approaches, including Magic3D \cite{lin2023magic3d}, Fantasia3D \cite{chen2023fantasia3d}, and ProlificDreamer \cite{wang2023prolificdreamer}. Notably, these competitors require multi-stage training, which is not needed in our model. This not only reduces our training cost but also maintains a simple training pipeline. Overall, our contributions can be summarized as follows. 

\begin{itemize}
    \item We provide an in-depth analysis of Score Distillation Sampling (SDS), the fundamental component in text-to-3D generation, and identify its key limitations for providing inconsistent and low-quality pseudo-GTs. This provides an explanation of the over-smoothing effect that exists in many approaches.  
    
    \item In response to SDS's limitations, we propose the Interval Score Matching (ISM). With invertible diffusion trajectories and interval-based matching, ISM significantly outperforms SDS with highly realistic and detailed results.
    
    \item By integrating with 3D Gaussian Splatting, our model achieves state-of-the-art performance, surpassing existing methods with less training costs. 
\end{itemize}

\section{Related Works}
\label{sec:text-3d-gen-background}
\vspace{0mm}\noindent\textbf{Text-to-3D Generation.} One work can be categorized as text-to-3D generation \cite{jain2022zero,poole2022dreamfusion,rombach2022high,saharia2022photorealistic,dhariwal2021diffusion,lin2023magic3d, metzer2023latent, chen2023fantasia3d, ho2022classifierfree,tang2023dreamgaussian,yi2023gaussiandreamer,shi2023MVD, armandpour2023re, chen2023gsgen}. 
As a pioneer, DreamField~\cite{jain2022zero} firstly train NeRF~\cite{mildenhall2021nerf} with CLIP~\cite{radford2021learning} guidance to achieve text-to-3D distillation. However, the results is unsatisfactory due to the weak supervision from CLIP loss.  With the advance of diffusion model, Dreamfusion~\cite{poole2022dreamfusion} introduces Score Distillation Sampling (SDS) to distill 3D assets from pre-trained 2D text-to-image diffusion models. 
SDS facilitates 3D distillation by seeking specific modes in a text-guide diffusion model, allowing for training a 3D model based on the 2D knowledge of diffusion models. This quickly motivates a great number of following works~\cite{poole2022dreamfusion, lin2023magic3d, chen2023fantasia3d, Zhang_Chen_Yang_Qu_Wang_Chen_Long_Zhu_Du_Zheng_2023, metzer2023latent, huang2023dreamwaltz,qian2023magic123} and becomes a critical integration of them. These works improve the performance of text-to-3D in various ways. 
For example, some of them~\cite{lin2023magic3d, metzer2023latent, chen2023fantasia3d, ho2022classifierfree,tang2023dreamgaussian,yi2023gaussiandreamer} improve the visual quality of text-to-3D distillation via modifying NeRF or introducing other advanced 3D representations. The other some~\cite{shi2023MVD, armandpour2023re, chen2023gsgen} focus on addressing the Janus problems, e.g., MVDream~\cite{shi2023MVD} propose to fine-tune the pre-trained diffusion models to make it 3D aware, and  GSGEN~\cite{chen2023gsgen} proposes a novel approach by introducing a 3D diffusion model for joint optimization. 
However, all these methods heavily rely on the Score Distillation Sampling. Albeit promising, SDS has shown over-smoothing effects in a lot of literatures~\cite{poole2022dreamfusion,lin2023magic3d,Zhang_Chen_Yang_Qu_Wang_Chen_Long_Zhu_Du_Zheng_2023,michel2022text2mesh}. Besides, it need coupling with a large conditional guidance scale~\cite{ho2022classifierfree}, leading to over-saturation results. There are also some very recent works \cite{wang2023prolificdreamer,zhu2023hifa,yu2023text,katzir2023noise} target at improving SDS. 
ProlificDreamer~\cite{wang2023prolificdreamer} proposes VSD to model 3D representation as a distribution. HiFA~\cite{zhu2023hifa} propose a iterative to estimate a better sampling direction. Although significant improve has been made, these works require a much longer training stage. 
CSD~\cite{yu2023text} and NFSD~\cite{katzir2023noise} are two concurrent works that analyze the components in the SDS to obtain empirical solutions to improve the original SDS. 
Our work is intrinsically different in the sense that it provides a systematic analysis on the the inconsistency and low-quality pseudo-ground-truths in SDS. And by introducing the Interval Score Matching, it achieves superior results without increasing the computational burden.  

\vspace{0mm}\noindent\textbf{Differentiable 3D Representations.} 
Differentiable 3D representation is a crucial integration of text-guided 3D generation. Given a 3D representation with trainable parameter $\theta$, a differentiable rendering equation $\vg(\theta,c)$ is used to render an image in camera pose $c$ of that 3D representation. As process is differentiable, we could train the 3D representation to fit our condition with backpropagation. Previously, various representations have been introduce to text-to-3D generations~\cite{mildenhall2021nerf,barron2021mip,verbin2022ref,ge2023ref,shen2021dmtet}. Among them,  NeRF~\cite{mildenhall2021nerf,lin2023magic3d,shi2023MVD} is the most common representation in text-to-3D generation tasks. 
The heavy rendering process of implicit representations makes it challenging for NeRF to produce high-resolution images that match the diffusion's resolution during distillation. Consequently, this limitation leads to suboptimal outcomes. To address this, textual meshes s~\cite{shen2021dmtet}, known for their efficient explicit rendering, are now used in this field to create detailed 3D assets \cite{lin2023magic3d,chen2023fantasia3d,wang2023prolificdreamer}, leading to better performance. Meanwhile, 3D Gaussian Splatting~\cite{kerbl20233d}, another effective explicit representation, demonstrates remarkable efficiency in reconstruction tasks. In this paper, we investigate 3D Gaussian Splatting~\cite{kerbl20233d} as the 3D representation in our framework. 

 \vspace{0mm}\noindent\textbf{Diffusion Models.} Another key component of text-to-3D generation is the diffusion model, which provides supervision for the 3D model. We briefly introduce it here to covers some notations. The Denoising Diffusion Probabilistic Model (DDPM)~\cite{ho2020denoising, song2021scorebased, saharia2022photorealistic} has been widely adopted for text-guided 2D image generation for its comprehensive capability. DDPMs assume $p(\vx_t|\vx_{t-1})$ as a diffusion process according to a predefined schedule $\beta_t$ on timestep $t$, that:
\begin{equation}
\label{eqn:addnoise}
\resizebox{0.6\linewidth}{!}{$p(\vx_t|\vx_{t-1}) = \mathcal{N}(\vx_t;\sqrt{1 - \beta_t}\vx_{t-1}, \beta_t \mI).
$}
\end{equation}
And the posterior $p_\phi(\vx_{t-1}|\vx_t)$ is modelled with a neural network $\phi$, where:
\begin{equation}
\label{eqn:denoise}
\resizebox{0.89\linewidth}{!}{$p_\phi(\vx_{t-1}|\vx_t) = \mathcal{N}(\vx_{t-1}; \sqrt{\bar\alpha_{t-1}}\mu_\phi(\vx_t), (1 - \bar\alpha_{t-1})\Sigma_\phi(\vx_t)),$}
\end{equation}
where $\bar\alpha_t \coloneqq (\prod_{1}^t  1 - \beta_t)$, and $\mu_\phi(\vx_t)$, $\Sigma_\phi(\vx_t)$ denote the predicted mean and variance given $x_t$, respectively. 

\section{Methodology}
\subsection{Revisiting the SDS}
\label{sec:sds_revisit}
As mentioned in Sec.~\ref{sec:text-3d-gen-background}, SDS~\cite{poole2022dreamfusion} pioneers text-to-3D generation by seeking modes for the conditional post prior in the DDPM latent space.
Denoting $\vx_0 \coloneqq \vg(\theta, c)$ as 2D views rendered from $\theta$, the posterior of noisy latent $x_t$ is defined as:
\begin{gather}
\label{eq:q(x)}
q^\theta(\vx_t) = \mathcal{N}(\vx_t;\sqrt{\bar\alpha_t}\vx_{0}, (1 - \bar\alpha_t)\mI).
\end{gather}
Meanwhile, SDS adopts pretrained DDPMs to model the conditional posterior of $p_\phi(\vx_t | y)$. Then, SDS aims to distill 3D representation $\theta$ via seeking modes for such conditional posterior, which can be achieved by minimizing the following KL divergence for all $t$:
\begin{equation}
\label{eq:SDS_kl}
\resizebox{0.91\linewidth}{!}{$\min_{\theta\in\Theta} \gL_{\mbox{\tiny SDS}}(\theta) \coloneqq \E_{t,c}\left[\omega(t) \kl{q^\theta(\vx_t)}{p_\phi(\vx_t|y)} \right].$}
\end{equation}
Further, by reusing the weighted denoising score matching objective~\cite{ho2020denoising,song2021scorebased} for DDPM training, the Eq.~\eqref{eq:SDS_kl} is reparameterized as:
\begin{equation}
\label{eq:sds_loss}
\resizebox{0.8\linewidth}{!}{$\min_{\theta\in\Theta} \gL_{\mbox{\tiny SDS}}(\theta) \coloneqq \E_{t, c} \left[\omega(t)||\vepsilon_\phi(\vx_t, t, y) - \vepsilon||^2_2\right],$}
\end{equation}
where $\vepsilon\sim\gN(\bm{0},\mI)$ is the ground truth denoising direction of $\vx_t$ in timestep $t$.
And the $\vepsilon_\phi(\vx_t, t, y)$ is the predicted denoising direction with given condition $y$. Ignoring the UNet Jacobian~\cite{poole2022dreamfusion}, the gradient of SDS loss on $\theta$ is given by:
\begin{equation}
\label{eq:sds_grad}
    \resizebox{0.75\linewidth}{!}{$\nabla_{\theta} \gL_{\mbox{\tiny SDS}}(\theta) \approx \E_{t,\vepsilon,c}\,[\omega(t) (\underbrace{\vepsilon_\phi(\vx_t,t,y) - \vepsilon}_{\text{SDS update direction}}) \frac{\partial\vg(\theta,c)}{\partial\theta}]\mbox{.}$}
\end{equation}

\vspace{0mm}\noindent\textbf{Analysis of SDS.} To lay a clearer foundation for the upcoming discussion, we denote $\gamma(t) = \frac{\sqrt{1 - \bar \alpha_t}}{\sqrt{\bar \alpha_t}}$ and equivalently transform Eq.~\eqref{eq:sds_loss} into an alternative form as follows:
\begin{equation}
\label{eq:sds_alter}
\resizebox{0.89\linewidth}{!}{$
\begin{aligned}
    \min_{\theta\in\Theta} \gL_{\mbox{\tiny SDS}}(\theta) &\coloneqq \E_{t,\vepsilon,c}\,\left[\frac{\omega(t)}{\gamma(t)}||\gamma(t)(\vepsilon_\phi(\vx_t,t,y)-\vepsilon) + \frac{(\vx_t - \vx_t)}{\sqrt{\bar \alpha_t}}||^2_2 \frac{\partial\vg(\theta,c)}{\partial\theta}\right] \\
    &= \E_{t,\vepsilon,c}\,\left[\frac{\omega(t)}{\gamma(t)}||\vx_0 - \hat{\vx}_0^t||^2_2 \frac{\partial\vg(\theta,c)}{\partial\theta}\right].
\end{aligned}
$}
\end{equation}
where $\vx_t \sim q^\theta(\vx_t)$ and $\hat{\vx}_0^t= \frac{\vx_t -\sqrt{1 - \bar \alpha_t}\vepsilon_\phi(\vx_t,t,y)}{\sqrt{\bar \alpha_t}}$. Consequently, we can also rewrite the gradient of SDS loss as:
\begin{equation}
\label{eq:sds_grad_alter}
    \resizebox{0.7\linewidth}{!}{$\nabla_{\theta} \gL_{\mbox{\tiny SDS}}(\theta) = \E_{t,\vepsilon,c}\,[\frac{\omega(t)}{\gamma(t)} (\vx_0-\hat{\vx}_0^t) \frac{\partial\vg(\theta,c)}{\partial\theta}]\mbox{.}$}
\end{equation}

In this sense, the SDS objective can be viewed as matching the view $\vx_0$ of the 3D model with $\hat{\vx}^t_0$ (i.e., the pseudo-GT) that DDPM estimates from $\vx_t$ in a single-step.
However, we have discovered that this distillation paradigm overlooks certain critical aspects of the DDPM.
In Fig.~\ref{fig:sds_drawbacks}, we show that the pretrained DDPM tends to predict feature-inconsistent pseudo-GTs, which are sometimes of low quality during the distillation process. However, all updating directions yielded by Eq.~\eqref{eq:sds_grad_alter} under such undesirable circumstances would be updated to the $\theta$, and inevitably lead to over-smoothed results. 
We conclude the reasons for such phenomena from two major aspects. 
First, it is important to note a key intuition of SDS: it generates pseudo-GTs with 2D DDPM by referencing the input view $\vx_0$. And afterward, SDS exploits such pseudo-GTs for $\vx_0$ optimization. 
As disclosed by Eq.~\eqref{eq:sds_grad_alter}, SDS achieves this goal by first perturbing $\vx_0$ to $\vx_t$ with random noises, then estimating $\hat{\vx}^t_0$ as the pseudo-GT.
However, we notice that the DDPM is very sensitive to its input, where minor fluctuations in $\vx_t$ would change the features of pseudo-GT significantly. Meanwhile, we find that not only the randomness in the noise component of $\vx_t$, but also the randomness in the camera pose of $\vx_0$ could contribute to such fluctuations, which is inevitable during the distillation. 
Optimizing $\vx_0$ towards inconsistent pseudo-GTs ultimately leads to feature-averaged outcomes, as depicted in the last column of Fig.~\ref{fig:sds_drawbacks}. 

Second, Eq.~\eqref{eq:sds_grad_alter} implies that SDS obtains such pseudo-GTs with a single-step prediction for all $t$, which neglects the limitation of single-step-DDPM that are usually incapable of producing high-quality results. As we also show in the middle columns of Fig.~\ref{fig:sds_drawbacks}, such single-step predicted pseudo-GTs are sometimes detail-less or blurry, which obviously hinders the distillation. 
Consequently, we believe that distilling 3D assets with the SDS objective might be less ideal. Motivated by such observations, we aim to settle the aforementioned issues in order to achieve better results.

\subsection{Interval Score Matching}
\label{sec:ISM}
Note that the aforementioned problems originate from the fact that $\hat{\vx}^t_0$, which serves as the \textit{pseudo-ground-truth} to match with $\vx_0=\vg(\theta,c)$, is inconsistent and sometimes low quality. 
In this section, we provide an alternative solution to SDS that significantly mitigates these problems. 

Our core idea lies in two folds. 
First, we seek to obtain more consistent pseudo-GTs during distillation, regardless of the randomness in noise and camera pose. Then, we generate such pseudo-GTs with high visual quality.

\vspace{0mm}\noindent\textbf{DDIM Inversion.}
As discussed above, we seek to produce more consistent pseudo-GTs that are aligned with $\vx_0$. Thus, instead of producing $\vx_t$ stochastically with Eq.~\eqref{eq:q(x)}, we employ the DDIM inversion to predict the noisy latent $\vx_t$. 
Specifically, DDIM inversion predicts a invertible noisy latent trajectory $\{\vx_{\delta_T}, \vx_{2\delta_T}, ..., \vx_t\}$ in an iterative manner:
\begin{equation} 
\label{eq:DDIM_inversion}
\begin{split}
    \vx_t = \sqrt{\bar \alpha_t}\hat{\vx}_0^{s} + \sqrt{1 - \bar \alpha_t }\vepsilon_\phi(\vx_s, s, \emptyset_{}) \\
    =  \sqrt{\bar \alpha_t}(\hat{\vx}_0^{s} + \gamma(t)\vepsilon_\phi(\vx_s, s, \emptyset_{})),
\end{split}
\end{equation}  
where $s = t - \delta_T$, and $\hat{x}_0^{s} = \frac{1}{\sqrt{\bar \alpha_s}}\vx_{s} - \gamma(s)\vepsilon_\phi(\vx_s, s, \emptyset_{})$. With some simple computation, we organize $\hat{\vx}_0^s$ as:
\begin{equation} 
\resizebox{0.85\linewidth}{!}{$
\begin{split}
    \hat{\vx}_0^s = \vx_0 - &\gamma(\delta_T)[\vepsilon_\phi(\vx_{\delta_T}, \delta_T, \emptyset) - \vepsilon_\phi(\vx_{0}, 0, \emptyset_{})] - \cdots
    \\
    - &\gamma(s)[\vepsilon_\phi(\vx_{s}, s, \emptyset) - \vepsilon_\phi(\vx_{s-\delta_T}, s-\delta_T, \emptyset_{})],
\end{split}
$}
\end{equation}
Thanks to the invertibility of DDIM inversion, we significantly increase the consistency of the pseudo-GT (i.e., the $\hat{\vx}_0^t$) with $\vx_0$ for all $t$, 
which is important for our subsequent operations.
To save space, please refer to our supplement for analysis.

\vspace{0mm}\noindent\textbf{Interval Score Matching.} 
Another limitation of SDS is that it generates pseudo-GTs with a single-step prediction from $x_t$ for all $t$, making it challenging to guarantee high-quality pseudo-GTs. 
On this basis, we further seek to improve the visual quality of the pseudo-GTs. 
Intuitively, this can be achieved by replacing the single-step estimated pseudo-GT $\hat{\vx}_0^t = \frac{1}{\sqrt{\bar \alpha_t}}\vx_{t} - \gamma(t)\vepsilon_\phi(\vx_t, t, y)$ with a multi-step one, denoted as $\tilde{\vx}_0^t \coloneqq \tilde{\vx}_0$, following the multi-step DDIM denoising process, i.e., iterating 
\begin{equation} 
\label{eq:DDIM}
    \tilde{\vx}_{t-\delta_T} = \sqrt{\bar \alpha_{t-\delta_T}}(\hat{\vx}_0^t + \gamma(t-\delta_T)\vepsilon_\phi(\vx_t, t, y))
\end{equation}
until $\tilde{\vx}_0$. Note that different from the DDIM inversion (Eq. \eqref{eq:DDIM_inversion}), this denoising process is conditioned on $y$. This matches the behavior of SDS (Eq. \eqref{eq:sds_grad}), i.e., SDS imposes unconditional noise $\vepsilon$ during forwarding and denoise the noisy latent with a conditional model $\vepsilon_\phi(\vx_t,t,y)$.

Intuitively, by replacing $\hat{\vx}_0^t$ in Eq.~\eqref{eq:sds_grad_alter} with $\tilde{\vx}_0^{t}$, we conclude a naive alternative of the SDS, where:
\begin{equation}
\label{eq:ism_grad_naive}
    \resizebox{0.6\linewidth}{!}{$\nabla_{\theta} \gL(\theta) = \E_{c}\,[\frac{\omega(t)}{\gamma(t)} (\vx_0-\tilde{\vx}_0^{t}) \frac{\partial\vg(\theta,c)}{\partial\theta}]\mbox{.}$}
\end{equation}
Although $\tilde{\vx}_0^{t}$ might produce higher quality guidance, it is overly time-consuming to compute, which greatly limits the practicality of such an algorithm. 
This motivates us to delve deeper into the problem and search for a more efficient approach.

Initially, we investigate the denoising process of $\tilde{\vx}_0^t$ jointly with the inversion process. 
We first unify the iterative process in Eq. \eqref{eq:DDIM} as 
\begin{equation}
    \label{eqn:ddim_denoise}
    \resizebox{0.89\linewidth}{!}{$
    \begin{split}
        \tilde{\vx}_0^{t} = \frac{\vx_t}{\sqrt{\bar \alpha_t}} - \gamma(t)\vepsilon_\phi(\vx_t, t, y) + \gamma(s)[\vepsilon_\phi(\vx_t, t, y) - \vepsilon_\phi(\tilde{\vx}_s, s, y)] &
        \\
        + \cdots + \gamma(\delta_T)[\vepsilon_\phi(\tilde{\vx}_{2\delta_T}, 2\delta_T, y) - \vepsilon_\phi(\tilde{\vx}_{\delta_T}, \delta_T, y)]&.
    \end{split}
    $}
\end{equation}
Then, combining Eq.~\eqref{eq:DDIM_inversion} with Eq.~\eqref{eqn:ddim_denoise}, we could transform Eq.~\eqref{eq:ism_grad_naive} as follows:
\begin{equation}
\label{eqn:x0-hatx0}
\resizebox{0.89\linewidth}{!}{$
    \begin{split}
    \resizebox{0.7\linewidth}{!}{$\nabla_{\theta} \gL(\theta) = \E_{t,c}\,[\frac{\omega(t)}{\gamma(t)} (\gamma(t)[\underbrace{\vepsilon_\phi(\vx_t, t, y) - \vepsilon_\phi(\vx_s, s, \emptyset)}_{\text{interval scores}}] + \eta_t) \frac{\partial\vg(\theta,c)}{\partial\theta}]\mbox{.}$}
    \end{split}
$}
\end{equation}
where we summarize the bias term $\eta_t$ as:
\begin{equation}
\resizebox{0.74\linewidth}{!}{$
    \begin{split}
         \eta_t = & + \gamma(s)[\vepsilon_\phi(\tilde{\vx}_s, s, y) - \vepsilon_\phi(\vx_{s-\delta_T}, s-\delta_T, \emptyset)]      \\
         & -\gamma(s)[\vepsilon_\phi(\vx_t, t, y) - \vepsilon_\phi(\vx_s, s, \emptyset)] 
         \\
         &+ ...
         \\
         & + \gamma(\delta_T)[\vepsilon_\phi(\tilde{\vx}_{\delta_T}, \delta_T, y) - \vepsilon_\phi(\vx_0, 0, \emptyset)] 
         \\
         & - \gamma(\delta_T)[\vepsilon_\phi(\tilde{\vx}_{2\delta_T}, 2\delta_T, y) - \vepsilon_\phi(\vx_{\delta_T}, \delta_T, \emptyset)].  
    \end{split}
$}
\end{equation}
Notably, $ \eta_t $ includes a series of neighboring interval scores with opposing scales, which are deemed to cancel each other out. Moreover, minimizing $\eta_t$ is beyond our intention since it contains a series of score residuals that are more related to $\delta_T$, which is a hyperparameter that is unrelated to 3D representation.
Thus, we propose to disregard $\eta_t$ to gain a boost in the training efficiency without compromising the distillation quality. Please refer to our supplement for more analysis and experiments about $\eta_t$.

\begin{figure}[t]
    \includegraphics[width=1.0\linewidth]{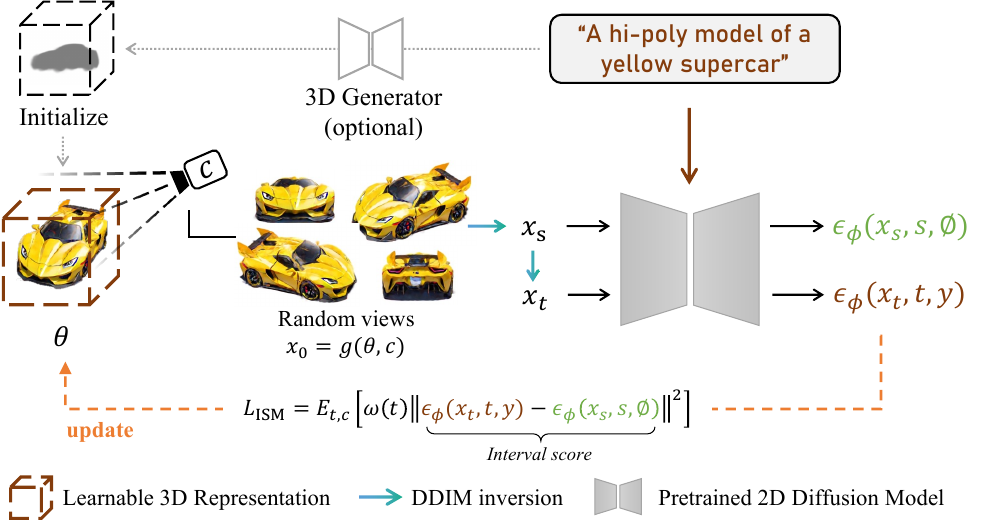}
    \vspace{-1.3em}
    \caption{\textbf{An overview of LucidDreamer.} In our paper, we first initialize the 3D representation (i.e. Gaussian Splatting~\cite{kerbl3Dgaussians}) $\theta$ via the pretrained text-to-3D generator~\cite{nichol2022point} with prompt $y$. Incorporate with pretrained 2D DDPM, we disturb random views $\vx_0 = \vg(\theta, c)$ to unconditional noisy latent trajectories $\{\vx_0, ..., \vx_s, \vx_t\}$ via DDIM inversion~\cite{song2022denoising}. Then, we update $\theta$ with the \textit{interval score}. Please refer to Sec.~\ref{sec:ISM} for details.}
    \label{fig:framework_overview}
\end{figure}

Consequently, we propose an efficient alternative to Eq.~\eqref{eq:ism_grad_naive} by disregarding the bias term $ \eta_t $ and focusing on minimizing the interval score, which we termed Interval Score Matching (ISM).
Specifically, with a given prompt $y$ and the noisy latents $\vx_s$ and $\vx_t$ generated through DDIM inversion from $x_0$, the ISM loss is defined as:
\begin{equation}
    \label{eq:native_ISM}
    \resizebox{0.89\linewidth}{!}{$\min_{\theta\in\Theta} \gL_{\mbox{\tiny ISM}}(\theta) \coloneqq \E_{t,c} \left[\omega(t)||\vepsilon_\phi(\vx_{t}, t, y) - \vepsilon_\phi(\vx_s, s, \emptyset_{})||^2\right].$}
\end{equation}
Following~\cite{poole2022dreamfusion}, the gradient of ISM loss over $\theta$ is given by:
\begin{equation}
\label{eq:ISM_grad}
\resizebox{0.85\linewidth}{!}{$\nabla_{\theta} \gL_{\mbox{\tiny ISM}}(\theta) \coloneqq \E_{t,c}\,[\omega(t) (\underbrace{\vepsilon_\phi(\vx_{t}, t, y_{}) - \vepsilon_\phi(\vx_s, s, \emptyset_{})}_{\text{ISM update direction}}) \frac{\partial\vg(\theta,c)}{\partial\theta}]\mbox{.}$}
\end{equation}
Despite omitting $ \eta_t $ from Equation~\eqref{eqn:x0-hatx0}, the core of optimizing the ISM objective still revolves around updating $ \vx_0 $ towards pseudo-GTs that are \textit{feature-consistent}, \textit{high-quality}, yet \textit{computationally friendly}.
Hence, ISM aligns with the fundamental principles of SDS-like objectives~\cite{poole2022dreamfusion, wang2023prolificdreamer, graikos2022diffusion} albeit in a more refined manner. 

As a result, ISM presents several advantages over previous methodologies. Firstly, owing to ISM providing consistent, high-quality pseudo-GTs, we produce high-fidelity distillation outcomes with rich details and fine structure, eliminating the necessity for a large conditional guidance scale~\cite{ho2022classifierfree} and enhancing the flexibility for 3D content creation. Secondly, unlike the other works~\cite{wang2023prolificdreamer,luo2023diffinstruct}, transitioning from SDS to ISM takes marginal computational overhead. Meanwhile, although ISM necessitates additional computation costs for DDIM inversion, it does not compromise the overall efficiency since 3D distillation with ISM usually converges in fewer iterations. Please refer to our supplement for more discussion. 

\begin{algorithm}[!t] 
    \caption{Interval Score Matching}
    \label{alg:ISM}
    \begin{spacing}{1.15}
            \begin{algorithmic}[1]
                \STATE Initialization: DDIM inversion step size $\delta_T$ and $\delta_S$, 
                     \\~~~~~~~~~~~~~~~~~~~~~~~the target prompt $y_{}$
                \WHILE{$\theta$ is not converged}
                    \STATE Sample: \resizebox{0.55\linewidth}{!}{$\vx_0 = g(\theta, c), t \sim \mathcal{U}(1, 1000)$}
                    \STATE let $s = t - \delta_T$ and $n = s / \delta_S$
                    \FOR{$i = [0, ..., n-1]$}
                        \STATE \resizebox{0.75\linewidth}{!}{$\hat{\vx}_0^{i \delta_S} = \frac{1}{\sqrt{\bar\alpha_{i \delta_S}}} (\vx_{i \delta_S} - \sqrt{1 - \bar\alpha_{i \delta_S}}\vepsilon_\phi(\vx_{i \delta_S}, i\delta_S, \emptyset_{}))$} 
                        
                        \STATE \resizebox{0.86\linewidth}{!}{$\vx_{(i+1)\delta_S} = \sqrt{\bar\alpha_{(i+1)\delta_S}}\hat{\vx}_0^{i\delta_S} + \sqrt{1 - \bar\alpha_{(i+1)\delta_S}}\vepsilon_\phi(\vx_{i\delta_S}, i\delta_S, \emptyset_{})$} 
                    \ENDFOR
                    \STATE predict $\vepsilon_\phi(\vx_s, s, \emptyset)$, then step $\vx_s \rightarrow \vx_t$ via\\
                           $\vx_t = \sqrt{\bar\alpha_{t}}\hat{\vx}_0^{s} + \sqrt{1 - \bar\alpha_{t}}\vepsilon_\phi(\vx_{s}, s, \emptyset_{})$
                    \STATE predict $\vepsilon_\phi(\vx_t, t, y)$ and compute ISM gradient \\
                           $\nabla_\theta L_{\text{ISM}} = \omega(t)(\vepsilon_\phi(\vx_{t}, t, y_{}) - \vepsilon_\phi(\vx_{s}, s, \emptyset_{}))$
                    \STATE update $\vx_0$ with $\nabla_\theta L_{\text{ISM}}$
                \ENDWHILE
            \end{algorithmic}
    \end{spacing}
\end{algorithm}

\begin{figure*}[!t]
    \hsize=\textwidth
    \centering
    \includegraphics[width=1.0\linewidth]{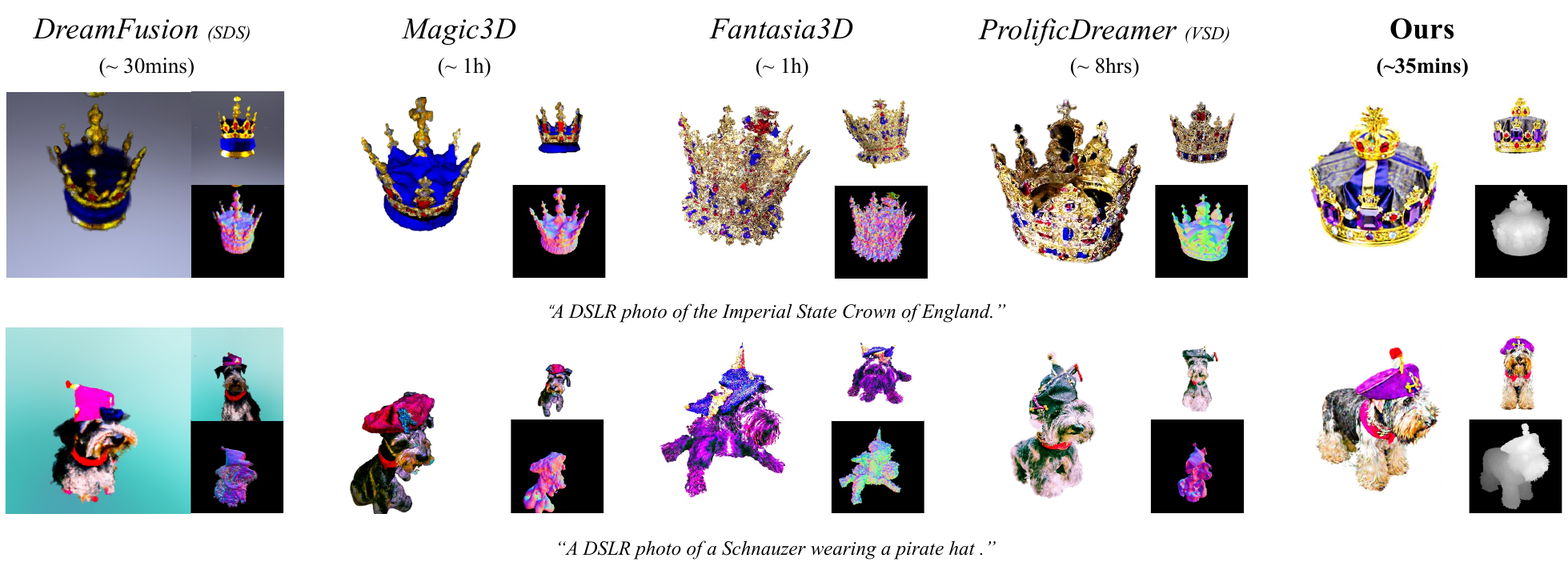}
    \vspace{-2em}
    \caption{\textbf{Comparison with baselines methods in text-to-3D generation.} Experiment shows that our approach is capable of creating 3D content that matches well with the input text prompts with high fidelity and intricate details. The running time of our method is measured on a single A100 GPU with a view batch size of 4, $\delta_S = 200$. Please zoom in for details.}
    \label{fig:compare}
\end{figure*}

Meanwhile, as the standard DDIM inversion usually adopts a fixed stride, it increases the cost for trajectory estimation linearly as $t$ goes larger. 
However, it is usually beneficial to supervise $\theta$ at larger timesteps. 
Thus, instead of estimating the latent trajectory with a uniform stride, we propose to accelerate the process by predicting $\vx_s$ with larger step sizes $\delta_S$. We find such a solution reduces the training time dramatically without compromising the distillation quality. In addition, we present a quantitative analysis of the impact of $\delta_T$ and $\delta_S$ in Sec.~\ref{sec:ablation}. 
Overall, we summarize our proposed ISM in Fig.~\ref{fig:framework_overview} and Algorithm~\ref{alg:ISM}. 

\subsection{The Advanced Generation Pipeline}
\label{sec:3DGS-pipeline}
We also explore the factors that would affect the visual quality of text-to-3D generation and propose an advanced pipeline with our ISM. Specifically, we introduce 3D Guassians Splatting (3DGS) as our 3D representation and 3D  point cloud generation models for initialization. 

\vspace{0mm}\noindent\textbf{3D Gaussian Splatting.} Empirical observations of existing works indicate that increasing the rendering resolution and batch size for training would significantly improve the visual quality. However, most learnable 3D representations that have been adopted in the text-to-3D generation~\cite{wang2023prolificdreamer, poole2022dreamfusion, shi2023MVD} are relatively time and memory-consuming. In contrast, 3D Gaussian Splatting~\cite{kerbl20233d} provides highly efficient in both rendering and optimizing. This drives our pipeline to achieve high-resolution rendering and large batch size even with more limited computational resources.

\vspace{0mm}\noindent\textbf{Initialization.} Most previous methods~\cite{poole2022dreamfusion, wang2023prolificdreamer, shi2023MVD, chen2023fantasia3d} usually initialize their 3D representation with limited geometries like box, sphere, and cylinder, which could lead to undesired results on non-axial-symmetric objects. Since we introduce the 3DGS as our 3D representation, we can naturally adopt several text-to-point generative models~\cite{nichol2022point} to generate the coarse initialization with humans prior. This initialization approach greatly improves the convergence speed, as shown in Sec.~\ref{sec:ablation}. 

\begin{figure}[t]
    \vspace{-2em}
    \includegraphics[width=1.0\linewidth]{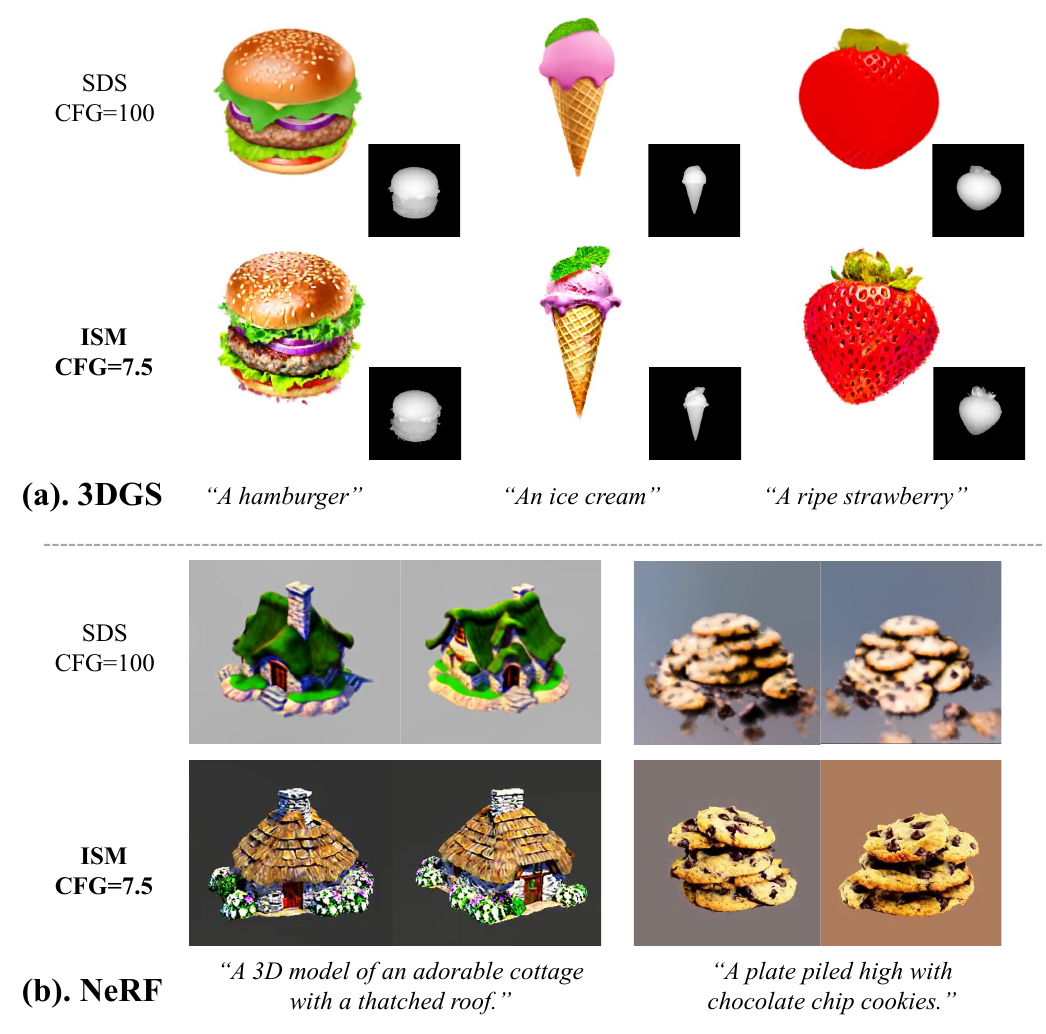}
    \vspace{-1.3em}
    \caption{\textbf{A comparison of SDS~\cite{poole2022dreamfusion} and ISM with different 3D models.} It shows that either using (a). 3DGS or (b). NeRF, the results of SDS tend to be smooth, whereas our ISM excels in distilling more realistic content and is rich in detail. Please zoom in for details.}
    \label{fig:sds_comparison}
    \vspace{-1em}
\end{figure}

\section{Experiments}
\label{sec:experiments}
\vspace{0mm}\noindent\textbf{Text-to-3D Generation.} We show the generated results of LucidDreamer in Fig.~\ref{fig:teaser} with original stable diffusion~\cite{rombach2022high} (below the dashed line) and various fintune checkpoints~\cite{Civitai_Lykon_2023,Civitai_Zovya_2023,Civitai_7whitefire7_2023}\footnote{Term of Service: https://civitai.com/content/tos} (above the dashed line). The results demonstrate that LucidDreamer is capable of generating 3D content that is highly consistent with the semantic cues of the input text. It excels in producing realistic and intricate appearances, avoiding issues of excessive smoothness or over-saturation, such as in the details of character portraits or hair textures. Furthermore, our framework is not only proficient in accurately generating common objects but also supports creative creations, like imagining unique concepts such as "Iron Man with white hair" (Fig. \ref{fig:teaser}). 

\vspace{0mm}\noindent\textbf{Generalizability of ISM.} 
To evaluate the generalizability of ISM, we conduct a comparison with ISM and SDS in both explicit representation (3DGS~\cite{kerbl3Dgaussians}) and implicit representation (NeRF~\cite{mildenhall2021nerf}). Notably, we follow the hyperparameter design of ProlificDreamer in the NeRF comparison. As shown in Fig~\ref{fig:sds_comparison}, our ISM provides fined-grained details even with normal CFG (7.5) in both NeRF~\cite{mildenhall2021nerf} and 3D Gaussian Splatting~\cite{kerbl3Dgaussians} (3DGS), which is significantly better than the SDS. This is a clear demonstration of the generalizability of our ISM.

\begin{figure}[!t]
    \includegraphics[width=1.0\linewidth]{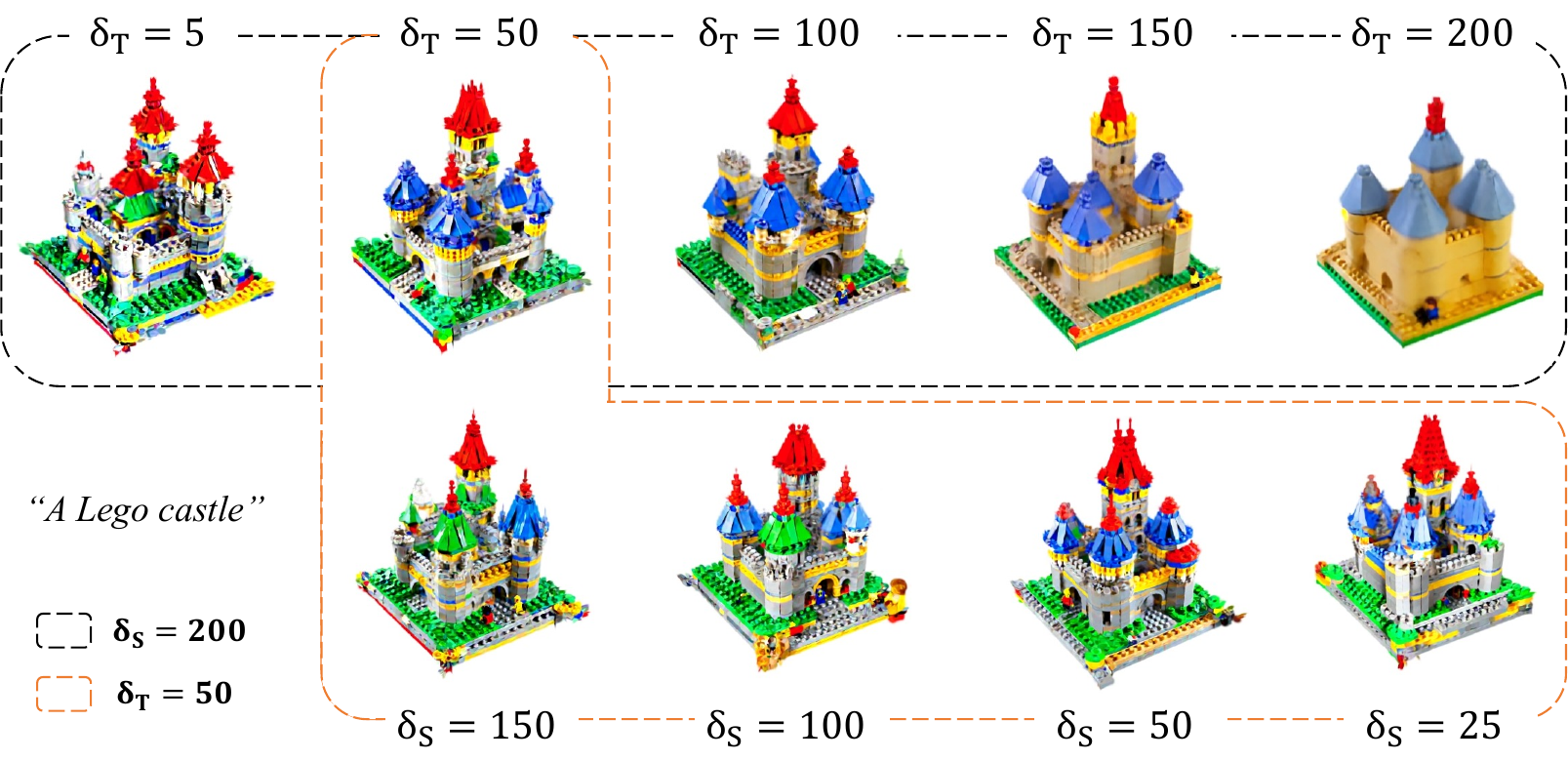}
    \vspace{-1.2em}
    \caption{\textbf{ISM with Different $\delta_T$ and $\delta_S$.} We fix $\delta_T = 50$ (orange dashed box) and $\delta_S = 200$ (black dashed box) respectively to compare the influence of these hyperparameters qualitatively. }
    \label{fig:interval_ablate}
    \vspace{-0.5em}
\end{figure}

\vspace{0mm}\noindent\textbf{Qualitative Comparison.} 
We compare our model with current SoTA baselines~\cite{wang2023prolificdreamer,chen2023fantasia3d,poole2022dreamfusion, lin2023magic3d} reimplemented by Three-studio~\cite{threestudio2023}. We all use the stable diffusion 2.1 for distillation and all experiments were conducted on A100 for fair comparison. As shown in Fig.~\ref{fig:compare}, our method achieves results regarding high fidelity and geometry consistency with less time and resource consumption. For example, the Crown generated by our framework exhibits more precise geometric structures and realistic colors, contrasting sharply with the geometric ambiguity prevalent in other baseline methods. Compared to Schnauzer generated by other methods, our approach produces Schnauzer with hair texture and overall body shape that is closer to reality, showing a clear advantage. Meanwhile, since the Point Generator introduces the geometry prior, the Janus problem is reduced in our framework.

\vspace{0mm}\noindent\textbf{User study.}
We conduct a user study to provide a comprehensive evaluation. Specifically, we select 28 prompts and generate objects using different Text-to-3D generation methods with each prompt. The users were asked to rank them based on the fidelity and the degree of alignment with the given text prompt. We show the average ranking to evaluate the users' preferences.
As shown in Tab.~\ref{tab:user_study}, our framework gets the highest average ranking in 6 selective methods. 
\begin{table}[h]    
    \centering
    \resizebox{1.0\linewidth}{!}{
    \begin{tabular}{|ccccc|c|}
        \hline
        DreamFusion \cite{poole2022dreamfusion} & Magic3D \cite{lin2023magic3d} & Text2Mesh\cite{michel2022text2mesh} & Fantasia3D \cite{chen2023fantasia3d} & ProlificDreamer \cite{wang2023prolificdreamer} & \textbf{Ours}
        \vspace{0.1em}
        \\
        \hline
        3.28 & 3.44 & 4.76 & 4.53 & 2.37 & \textbf{1.25}
        \\
        \hline
    \end{tabular}}
    \caption{We survey the users' preference ranking (\textbf{the smaller, the better}) averaged on 28 sets of text-to-3D generation results produced by baselines and our method, respectively. Our result is preferred by most users.}
    \label{tab:user_study}
\end{table}
Indicate that users consistently favored the 3D models generated by our framework. Please refer to our supplement for more details of the user study and more visual results.

\begin{figure}[t]
    \centering
    \vspace{-2em}
    \includegraphics[width=0.85\linewidth]{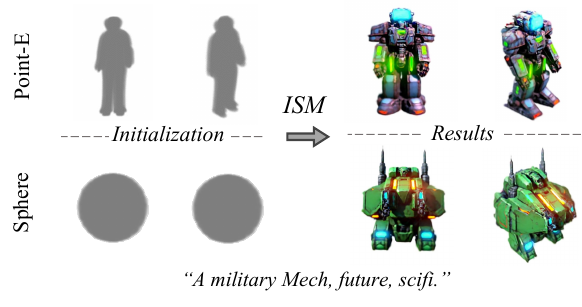}
    \vspace{-0.5em}
    \caption{\textbf{LucidDreamer with Different initialization.} We compare the results of two different initializations to evaluate the effectiveness of the Point Generator in our advanced pipeline.}
    \label{fig:pg_ablate}
    \vspace{-1em}
\end{figure}

\begin{figure*}[t]
    \hsize=\textwidth
    \centering
    \vspace{-2em}
    \includegraphics[width=1.0\linewidth]{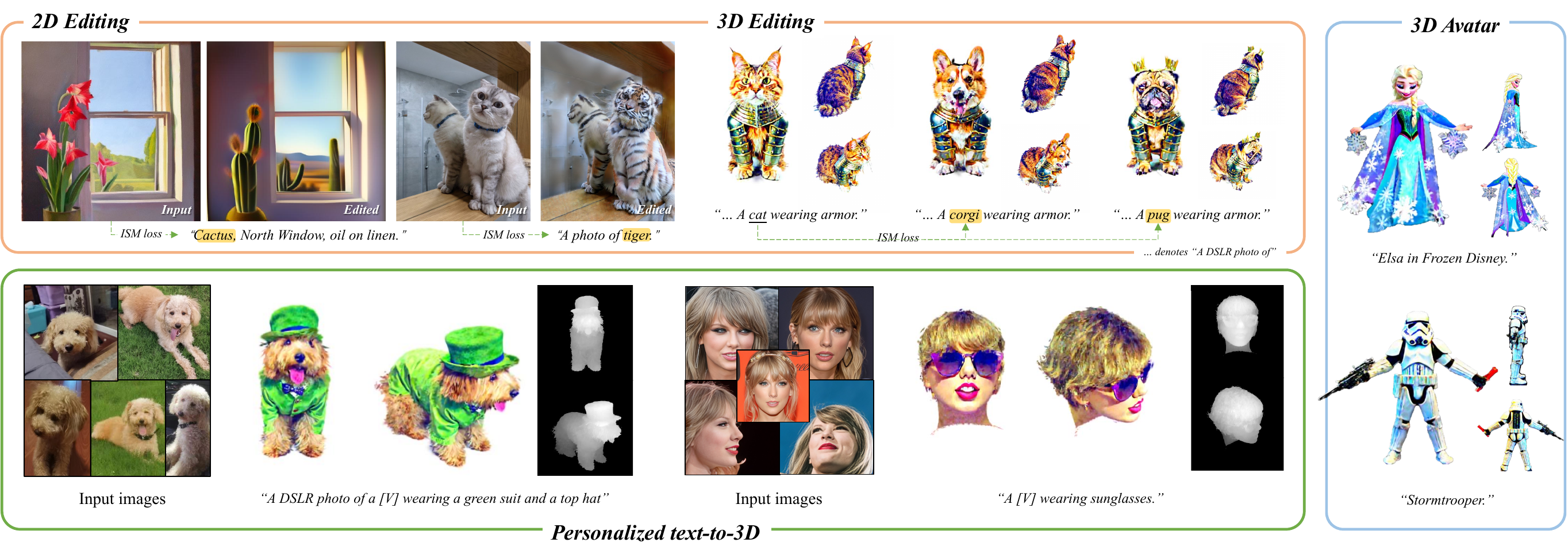}
    \vspace{-1.7em}
    \caption{\textbf{Applications of ISM.} We explore several applications with our proposed ISM, including the \textit{zero-shot 2D and 3D editing} (top left), \textit{personalized text-to-3D generation} with LoRA (bottom left), and \textit{3D avatar generation}. Generally, our proposed ISM as well as the Advanced 3D generation pipeline performs surprisingly well across various tasks. Please refer to our paper for more details.}
    \vspace{-4mm}
    \label{fig:applications_ism}
\end{figure*}

\subsection{Ablation Studies}
\label{sec:ablation}

\vspace{0mm}\noindent\textbf{Effect of Interval Length.} We explore the effect of interval length $\delta_T$ and $\delta_S$ during training in this section. 
In Fig.~\ref{fig:interval_ablate}, we visualize the influence of $\delta_T$ and $\delta_S$. 
For a fixed $\delta_T$, an increasing $\delta_S$ takes marginal influence in the results but significantly saves the computational costs of DDIM inversion. 
Meanwhile, as the parameter $\delta_T$ increases, the results adopt a more natural color and simpler structure. 
However, this comes at the expense of detail. Thus, we conclude a trade-off in the selection of $\delta_T$. For instance, at higher $\delta_T$, castle walls appear smoother. Conversely, lower $\delta_T$ values enhance detail but can result in unnecessary visual anomalies, such as overly saturated color and the illusion of floating artifacts atop castle towers.
We hypothesize such observation is caused by the gradients provided by small intervals containing more detailed features but less structural supervision. Thus, we propose annealing the interval with the intuitive process of initially constructing the overall structures and subsequently incorporating fine-grained features. Moreover, this hyperparameter allows the user to generate objects with different levels of smoothness according to their preferences.

\vspace{0mm}\noindent\textbf{Initialization with Point Generators} We ablate the Point Generators in this section. Specifically, we train two 3D Gaussians from a random initialization and starting from a generated raw point cloud with a given prompt, respectively. 
In Fig.~\ref{fig:pg_ablate}, we compare the distillation results with the same prompts but different. With the parameter and random seed guaranteed to be constant, 3D Gaussian with point initialization has a better result in geometry.

\section{Applications}
This section further explores the applications of LucidDreamer. Specifically, we combine our framework with advanced conditioning techniques and achieve some real-world applications.

\vspace{0mm}\noindent\textbf{Zero-shot Avatar Generation.}
We expand our framework to produce pose-specific avatars by employing the Skinned Multi-Person Linear Model (SMPL)~\cite{loper2023smpl} as a geometry prior to initializing the 3D Gaussian point cloud. Then, we rely on ControlNet~\cite{zhang2023adding} conditioned on DensePose~\cite{guler2018densepose} signals to offer more robust supervision. Specifically, we render the 3D human mesh into a 2D image using pytorch3d based on sampled camera parameters and subsequently input it into the pre-trained DensePose model to acquire the human body part segmentation map as a DensePose condition. A more detailed framework is shown in the supplement. Following such an advanced control signal, we can achieve a high-fidelity avatar as shown in Fig.~\ref{fig:applications_ism}.

\vspace{0mm}\noindent\textbf{Personalized Text-to-3D.}
We also combine our framework with personalized techniques, LoRA~\cite{hu2021lora}. Using such techniques, our model can learn to tie the subjects or styles to an identifier string and generate images of the subjects or styles. For text-to-3D generation, we can use the identifier string for 3D generation of specific subjects and styles. As shown in Fig.~\ref{fig:applications_ism}, our method can generate personalized humans or things with fine-grained details. This also shows the great potential of our method in controllable text-to-3D generation by combining it with advanced personalized techniques.

\vspace{0mm}\noindent\textbf{Zero-shot 2D and 3D Editing.}
While our framework is primarily designed for text-to-3D generation tasks, extending ISM to editing is feasible due to the similarities in both tasks. Effortlessly, we can edit a 2D image or 3D representation in a conditional distillation manner, as ISM provides consistent update directions based on the input image, guiding it towards the target condition, as demonstrated in Fig.~\ref{fig:applications_ism}. Owing to space limitations, we reserve further customization of ISM for 2D/3D editing tasks for future exploration.

\section{Conclusions}
In this paper, we have presented a comprehensive analysis of the over-smoothing effect inherent in Score Distillation Sampling (SDS), identifying its root cause in the inconsistency and low quality of pseudo ground truth. Addressing this issue, we introduced Interval Score Matching (ISM), a novel approach that offers consistent and reliable guidance. Our findings demonstrate that ISM effectively overcomes the over-smoothing challenge, yielding highly detailed results without extra computational costs. Notably, ISM's compatibility extends to various applications, including NeRF and 3D Gaussian Splatting for 3D generation and editing, as well as 2D editing tasks, showcasing its exceptional versatility. Building upon this, we have developed \textit{LucidDreamer}, a framework that combines ISM with 3D Gaussian Splatting. Through extensive experimentation, we established that \textit{LucidDreamer} significantly surpasses current state-of-the-art methodologies. Its superior performance paves the way for a broad spectrum of practical applications, ranging from text-to-3D generation and editing to zero-shot avatar creation and personalized Text-to-3D conversions, among others.

\section{Appendix}

\subsection{Implementation details}
In our LucidDreamer framework, we adopt an explicit 3D representation, the 3D Gaussian Splatting (3DGS)~\cite{kerbl20233d}, for 3D distillation with our proposed Interval Score Matching (ISM) objective. 
To optimize 3DGS towards the pseudo-ground-truth (pseudo-GT) generated by diffusion models, we follow most training hyperparameters from the original 3DGS paper. 
Specifically, we implement a strategy of densifying and pruning the Gaussian at every 300 iteration interval until a total of 3000 iterations. As our ISM provides precise gradients, we observe a significantly high coverage speed. 
Consequently, we streamline our training process to consist of around 5000 iterations, substantially less than the original 10,000 iterations required in previous works~\cite{poole2022dreamfusion}. 
In terms of the initialization of 3DGS, we utilize the pretrained Point-E~\cite{nichol2022point} checkpoint. Also, for some asymmetrical objects, we adopt camera-dependent prompts during the training following Perp-Neg~\cite{armandpour2023re} to reduce the Janus problems further. 

\paragraph{LucidDreamer with negative prompts} Also, we find that negative prompts would further improve the generation quality, thus, we use the negative prompts from~\cite{katzir2023noise} in some cases. Denoting $y$ and $y_n$ as the positive and negative prompts, we predict the text-conditional score of the noisy latent $x_t$ following the classifier-free guidance~\cite{ho2022classifierfree}:
\begin{equation}
    \label{eqn:cfg}
    \resizebox{0.89\linewidth}{!}{$\vepsilon_\phi(x_t, t, y) = \vepsilon_\phi(x_t, t, y_n) + gs * (\vepsilon_\phi(x_t, t, y) - \vepsilon_\phi(x_t, t, y_n)),$}
\end{equation}
where $gs$ is the guidance scale of prompt $y$.


\begin{figure}[t]
    \centering
    \includegraphics[width=1.0\linewidth]{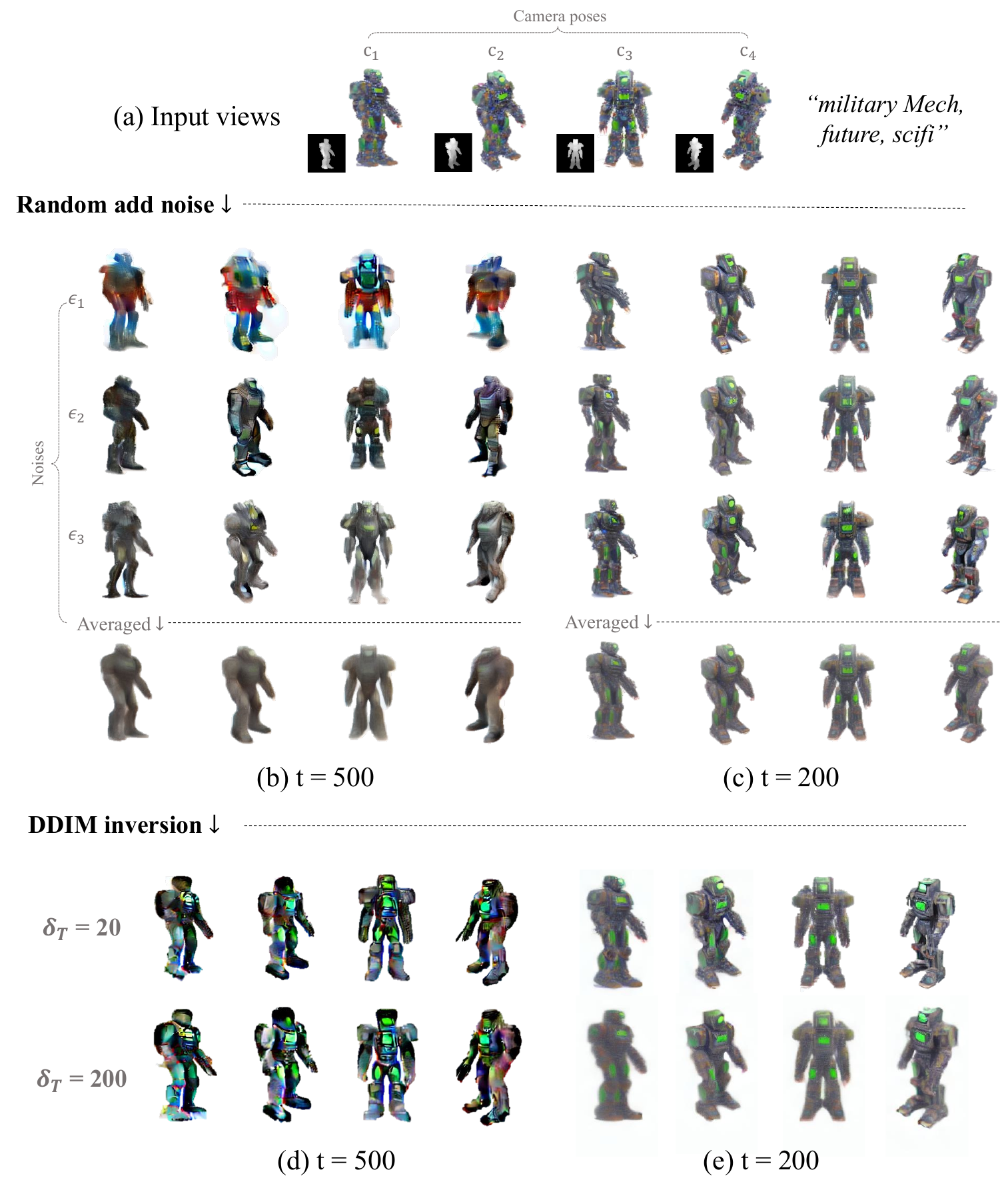}
    \vspace{-2em}
    \caption{\textbf{(a)}: The rendered $x_0$ from 3D representation with camera poses $c = \{c_1, ..., c_4\}$. \textbf{(b) and (c)}: pseudo-GTs $\hat{x}_0^t$ generated via randomly add noise $\vepsilon = \{\vepsilon_1, ... \vepsilon_3\}$ to $x_0$ at timestep $t = \{500, 200\}$. \textbf{(e) and (f)}: pseudo-GTs $\hat{x}_0^t$ generated via DDIM inversion with step size of $\delta_T = \{20, 200\}$ at timestep $t = \{500, 200\}$. Please zoom in for details.}
    \label{fig:inconsistent-pgt}
\end{figure}

\subsection{Inconsistency in SDS pseudo-GT}
\label{sec:SDS_inconsistency}
In our main paper, we discussed the inconsistency issue regards the pseudo-GTs produced by SDS~\cite{poole2022dreamfusion} in our revisiting of SDS. Specifically, it raised our concerns when we spotted significant inconsistency among the pseudo-GTs. Our investigation points out that such inconsistency is mainly caused by the following properties of the SDS algorithm: (1) randomness in timestep $t$; (2) randomness in the noise component $\vepsilon$ of $x_t$; (3) randomness in camera pose $c$.

To better explain the issue, we conducted a quantitative experiment on the inconsistency of pseudo-GTs with the aforementioned properties. In Fig.~\ref{fig:inconsistent-pgt} (a), we visualize the input views of 4 camera poses and the pseudo-GTs produced by SDS at different timesteps (Fig.~\ref{fig:inconsistent-pgt} (b) and (c)) and with different noise $\vepsilon$ (row 2 to 3). It can be seen that even with the noise fixed, the SDS pseudo-GTs tend to be inconsistent over different camera poses and timesteps and eventually lead to feature-averaged results, which is inevitable under the SDS distillation scheme.

\subsection{Complementary Experiments of ISM}
\subsubsection{Benefits of DDIM inversion}
In the previous section, we visualize the inconsistency issue of SDS pseudo-GTs. In the methodology section of our main paper, we propose to mitigate such a problem by introducing DDIM inversion for noisy latent estimation. Hence, we further examine the effect of replacing the vanilla add noise function for $x_0 \rightarrow x_t$ with DDIM inversion in Fig.~\ref{fig:inconsistent-pgt} (d) and (e). It can be seen that, the pseudo-GTs that incorporate with DDIM inversion are more similar to the input views in Fig.~\ref{fig:inconsistent-pgt} (a). Therefore, they are significantly more consistent feature and style-wise between different views and timesteps compared to Fig.~\ref{fig:inconsistent-pgt} (b) and (c).  Meanwhile, such a property holds when we increase $\delta_T$ from 20 to 200. 
Notably, DDIM inversion doesn't necessarily handle the quality problem of the pseudo-GTs generated with a single-step prediction with diffusion models. We will delve deeper into this problem in Sec.~\ref{sec:eta_t}.

\paragraph{3D distillation v.s. image-to-image translation} 
As we discussed in the main paper, ISM follows the basic intuition of SDS which generates pseudo-GTs with 2D diffusion models by referencing $x_0$. Intuitively, such a process is quite similar to the diffusion-based image-to-image translation tasks that have been discussed in some previous works~\cite{meng2021sdedit, su2022dual} that intend to alter the input image towards the given condition in a similar manner. 
In such a perspective, since SDS perturbs the clean sample $x_0$ with random noises, it encounters the same problem with SDEdit~\cite{meng2021sdedit} that it struggles to find an ideal timestep $t$ which ensures both the editability of the algorithm while maintaining the basic structure of the input image. 

Instead, our ISM adopts DDIM inversion to estimate $x_t$ from $x_0$ and thus share more common senses with DDIB~\cite{su2022dual} which mitigates the aforementioned problem. In essence, the DDIB proposes to edit images in a first ``DDIM inversion'' then ``DDIM denoising'' paradigm, which can be viewed as building two concatenated Schrödinger bridges~\cite{chen2021likelihood} that are intrinsically entropy-regularized optimal transport.
Similarly, our proposed ISM can be seen as first bridging the distribution of rendered images~$q(x_0)$ to the latent space~$p_\phi(x_t)$ of pretrained diffusion models $\phi$ via DDIM inversion, then, we bridge ~$p_\phi(x_t)$ to the target distribution~($p_\phi(x_0|y)$) via DDIM denoising. Then, we optimize $q(x_0)$ towards $p_\phi(x_0|y)$ along these bridges, which makes our ISM also an entropy-regularized optimal transport objective that is discussed in DDIB~\cite{su2022dual}. 
Consequently, our ISM is able to provide better pseudo-GTs for 3D distillation, which elucidates its superior performance over SDS.

\subsubsection{Discussion of $\eta_t$}
\label{sec:eta_t}
\begin{figure}[t]
    \centering
    \includegraphics[width=1\linewidth]{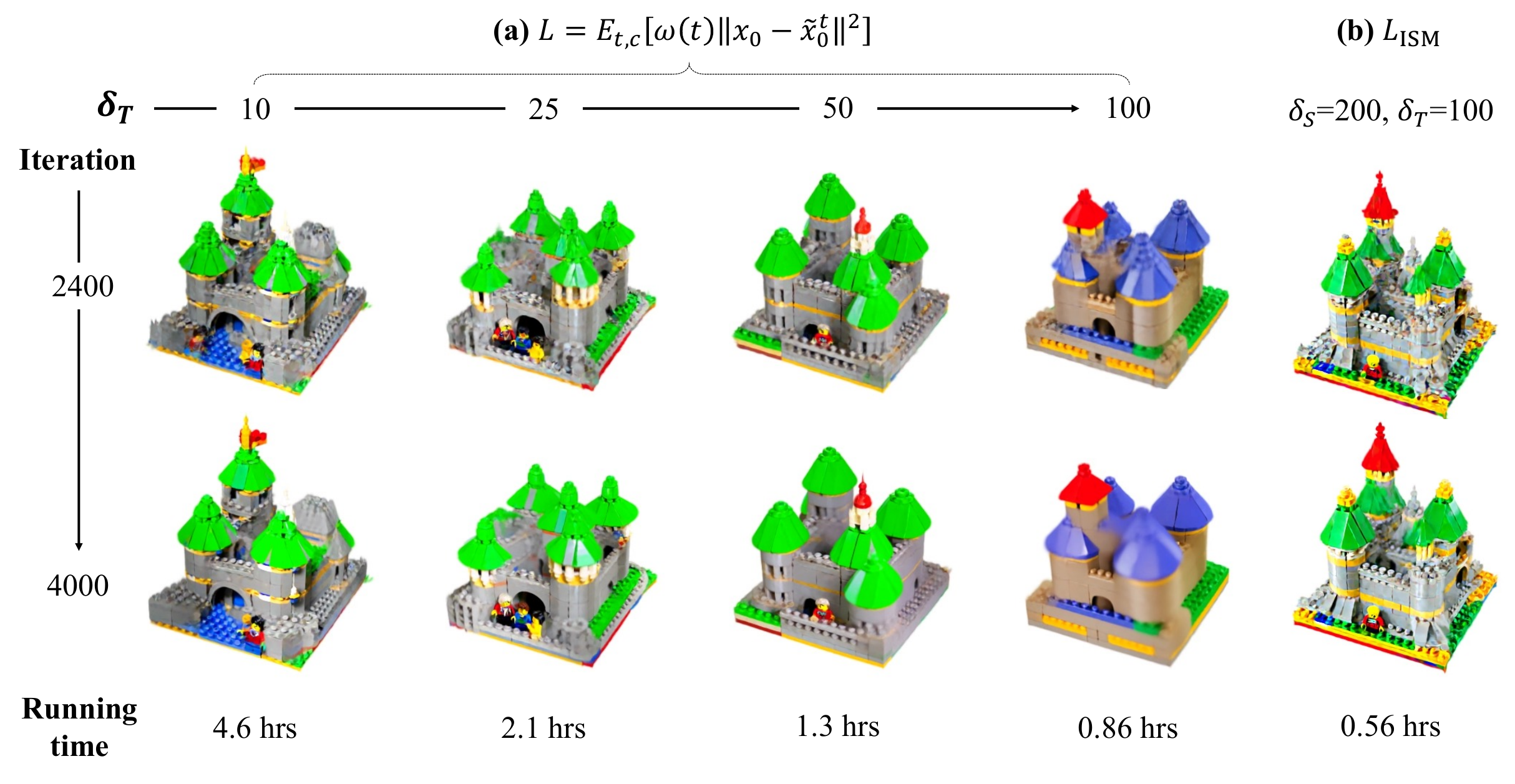}
    \vspace{-2em}
    \caption{\textbf{Comparison of the distillation results and running time.} (a) Distillation results with the naive objective (Eq.~\eqref{eqn:x0-hatx0}) at different $\delta_T = \{10, 25, 50, 100\}$. (b) Distillation results with our proposed ISM objective (Eq.~\eqref{eqn:ism}). Please zoom in for details.}
    \label{fig:denoise_x0}
    \vspace{-1em}
\end{figure}
\begin{figure*}[t]
    \centering
    \includegraphics[width=1.0\linewidth]{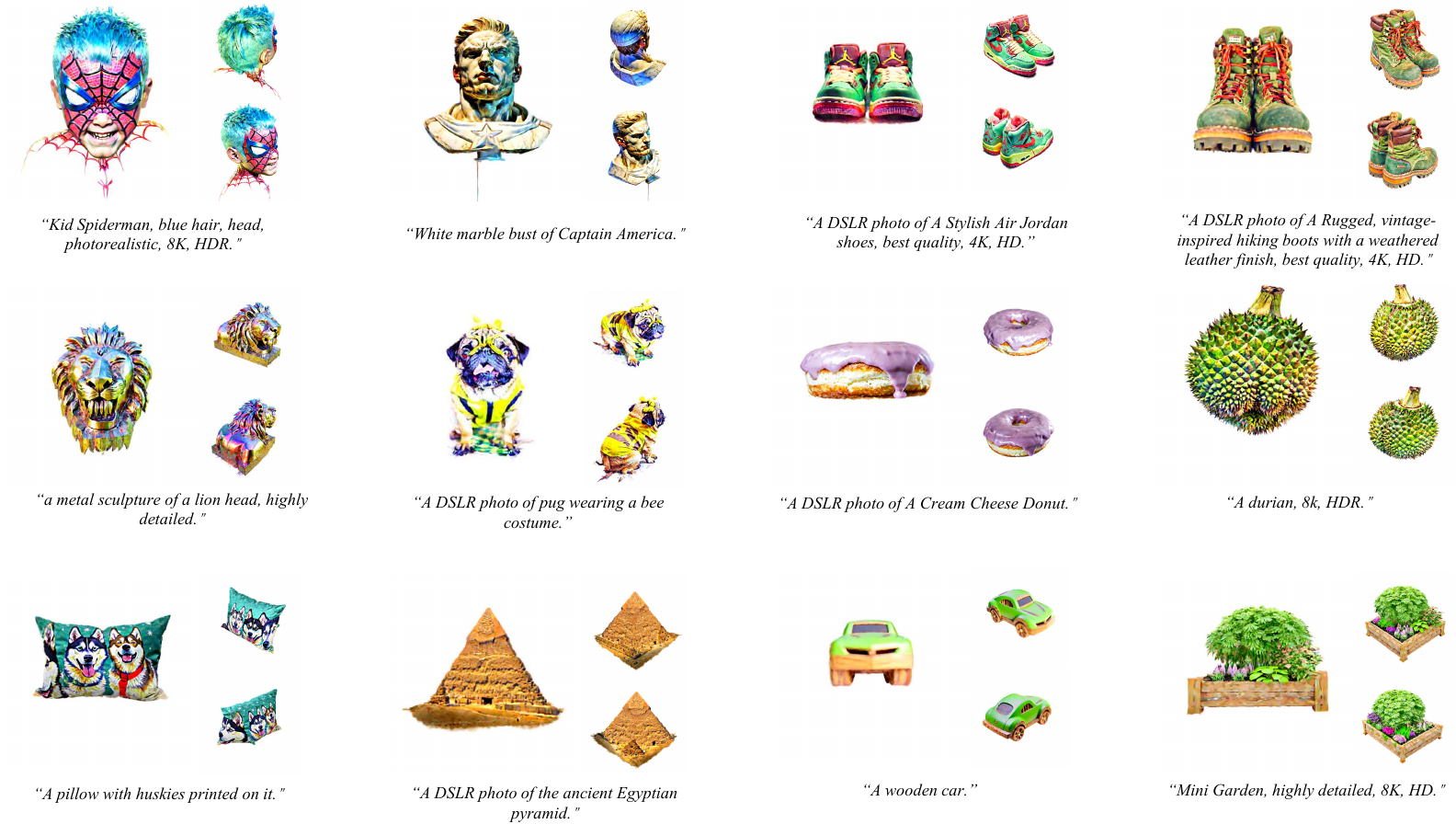}
    \caption{More results generated by our LucidDreamer framework. Please zoom in for details.}
    \label{fig:add_compare}
    \vspace{-1em}
\end{figure*}
\begin{figure}[t]
    \centering
    \includegraphics[width=1.0\linewidth]{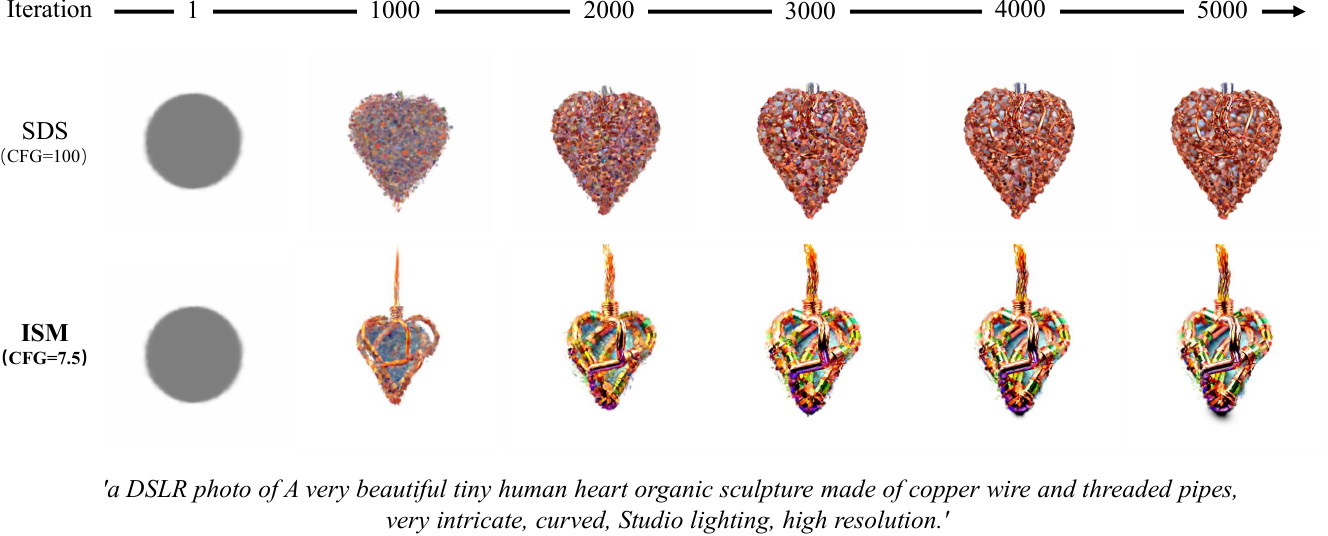}
    \caption{\textbf{Comparision of convergence speed.} Our ISM could quickly generate a clear structure (1000 iterations). While SDS failed. Please zoom in for details. }
    \label{fig:conv_speed}
    \vspace{-1em}
\end{figure}
\begin{figure}[t]
    \centering
    \includegraphics[width=0.95\linewidth]{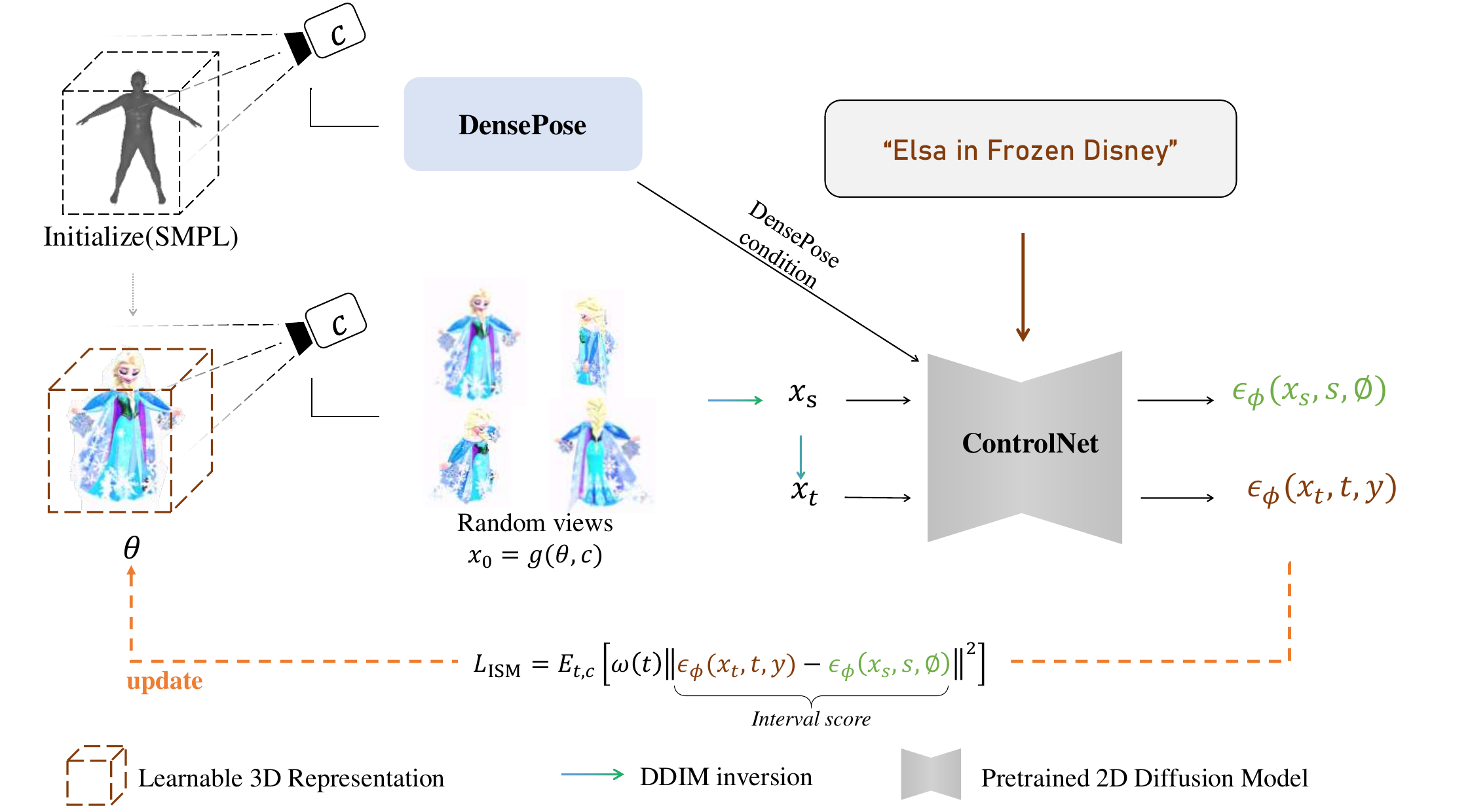}
    \caption{Framework of zero-shot Avatar Generation. In our paper, we first initialize the 3D representation via SMPL~\cite{loper2023smpl}. Then, we rely on ControlNet~\cite{zhang2023adding} conditioned on DensePose~\cite{guler2018densepose} signals provied by a pretrained DensePose predictor to offer more robust supervision.}
    \label{fig:avatar_pipeline}
    \vspace{-1em}
\end{figure}

In our main paper, we propose to replace the single-step pseudo-GT estimation adopted in SDS with a multi-step denoising operation. Then, combining the multi-step DDIM inversion with DDIM denoising with the same step size, we formulate our naive objective of 3D distillation as follows:
\begin{equation}
\label{eqn:x0-hatx0}
\resizebox{0.89\linewidth}{!}{$
    \begin{split}
    \gL(\theta) = &\E_{c}\,[\frac{\omega(t)}{\gamma(t)} ||\vx_0-\tilde{\vx}_0^{t}||^2]
    \\
    =&\E_{t,c}\,[\frac{\omega(t)}{\gamma(t)} ||\gamma(t)[\underbrace{\vepsilon_\phi(\vx_t, t, y) - \vepsilon_\phi(\vx_s, s, \emptyset)}_{\text{interval scores}}] + \eta_t||^2],
    \end{split}
$}
\end{equation}
where $\eta_t$ is a bias term depending on the denoising process $x_t \xrightarrow{} \Tilde{x}_0^t$. For example, when we adopt the step size of the DDIM inversion process $x_0 \xrightarrow{} x_t$, $\delta_T$, as the step size of the denoising process, it leads to:
\begin{equation}
\label{eqn:eta_t}
\resizebox{0.89\linewidth}{!}{$
\begin{split}
     \eta_t = & + \gamma(s)[\vepsilon_\phi(\tilde{\vx}_s, s, y) - \vepsilon_\phi(\vx_{s-\delta_T}, s-\delta_T, \emptyset)]      \\
     & -\gamma(s)[\vepsilon_\phi(\vx_t, t, y) - \vepsilon_\phi(\vx_s, s, \emptyset)] 
     \\
     & + \gamma(s -\delta_T)[\vepsilon_\phi(\tilde{\vx}_{s - \delta_T}, s - \delta_T, y) - \vepsilon_\phi(\vx_{s - 2\delta_T}, s - 2\delta_T, \emptyset)] 
     \\
     & - \gamma(s - \delta_T)[\vepsilon_\phi(\tilde{\vx}_{s}, s, y) - \vepsilon_\phi(\vx_{s - \delta_T}, s -\delta_T, \emptyset)]
     \\
     & + ...
     \\
     & + \gamma(\delta_T)[\vepsilon_\phi(\tilde{\vx}_{\delta_T}, \delta_T, y) - \vepsilon_\phi(\vx_0, 0, \emptyset)] 
     \\
     & - \gamma(\delta_T)[\vepsilon_\phi(\tilde{\vx}_{2\delta_T}, 2\delta_T, y) - \vepsilon_\phi(\vx_{\delta_T}, \delta_T, \emptyset)].  
\end{split}
$}
\end{equation}
Despite $\eta_t$ containing a series of neighboring interval scores with opposite scales that are deemed to cancel each other out, it inevitably leaks interval scores such as \resizebox{0.8\linewidth}{!}{$(\gamma(s) - \gamma(s - \delta_T))[\vepsilon_\phi(\tilde{\vx}_s, s, y) - \vepsilon_\phi(\vx_{s-\delta_T}, s-\delta_T, \emptyset)]$} and etc depending on the hyperparameters. 

Recap that the intuition behind Eq.~\eqref{eqn:x0-hatx0} is to distill update directions from all timestep $t$. Intuitively, because our algorithm would traverse all $t$, it is beyond our intention to distill update directions of the other timesteps (i.e., $s, s-\delta_T, ..., \delta_T$) when we focus on $t$. Furthermore, it is rather time-consuming to compute $\Tilde{x}_0^t$ since it requires equivalent steps of estimation for inversion and denoising.

In this paper, we propose to omit $\eta_t$ from Eq.~\eqref{eqn:x0-hatx0}, which leads to our ISM objective, where: 
\begin{equation}
\label{eqn:ism}
\resizebox{0.89\linewidth}{!}{$
    \begin{split}
        \gL_{\text{ISM}}(\theta) = \E_{t,c}\,[\omega(t)||\vepsilon_\phi(\vx_t, t, y) - \vepsilon_\phi(\vx_s, s, \emptyset)||^2].
    \end{split}
$}
\end{equation}
In Fig.~\ref{fig:denoise_x0}, we compare the distillation results of the naive objective versus ISM (with accelerated DDIM inversion). The results indicate that distilling 3D objects with ISM, as opposed to using the naive \eqref{eqn:x0-hatx0}, is not only markedly more efficient but also yields results with enhanced details. While the efficiency gain of ISM is anticipated, our hypothesis is that the observed improvement in details stems from the ISM objective's emphasis on updating directions solely at timestep $t$. This focus helps avoid the potentially inconsistent update directions at other timesteps $s, s-\delta_T, ..., \delta_T$ while we are not focusing on these timesteps. We will leave the investigation of such a problem to our future work.

\subsubsection{The convergence speed of ISM v.s. SDS}
We also compare the convergence speed of ISM and SDS. Specifically, we fixed the noise and hyperparameters and generated 3D assets using SDS and ISM, respectively. As shown in Fig.~\ref{fig:conv_speed}, our proposal (ISM) converges faster than SDS. \textit{e.g.} Our ISM generates a clear and reasonable structure using only 1000 iterations, while SDS is quite noisy at the same stage.

\subsection{Zero-shot Avatar Generation}
Our framework is highly adaptable to pose-specific avatar generation scenarios, as depicted in Fig~\ref{fig:avatar_pipeline}, which showcases the detailed workflow. To begin with, we utilize SMPL as an initialization step for positioning the Gaussian point cloud. Subsequently, we employ a pre-trained DensePose model to generate a segmentation map of the human body. This segmentation map serves as a conditional input for the pre-trained ControlNet, where we use an open-source controlnet-seg~\cite{zhang2023adding}.

\subsection{Details of User Study}
In this paper, we conduct a user study to research the
user’s preferences on the current SoTA text-to-3D methods.  In the user study, we ask the participants to compare the $360^{\circ}$ rendered video of generated assets from 6 different methods (including our proposal). We provide 28 sets of videos generated by different prompts. We collected 50 questionnaires from the internet and summarized the users' preferences, as shown in the main paper.

\subsection{More visual results}
We show additional generated results in Fig.~\ref{fig:add_compare}. It can be seen that our LucidDreamer could generate 3D assets with high visual quality and 3D consistency.

{\small
\bibliographystyle{ieeenat_fullname}
\bibliography{egbib}
}

\end{document}